\documentclass[11pt,a4paper]{article}
\usepackage[hyperref]{acl2018}
\usepackage{times}
\usepackage{latexsym}
\usepackage{amsmath}
\usepackage{CJK}
\usepackage{subfigure}
\usepackage{natbib}
\usepackage{pgfplots}
\usepackage{pgfplotstable}
\pgfplotsset{width=7.0cm,compat=1.5}
\usepackage{color,xcolor,framed}
\usepackage{url}

\aclfinalcopy 
\def\aclpaperid{702} 

\newcommand*{\affaddr}[1]{#1} 

\newcommand*{\email}[1]{\texttt{#1}}

\title{Global Encoding for Abstractive Summarization}

\author{Junyang Lin, Xu Sun, Shuming Ma, Qi Su\\
\affaddr{MOE Key Lab of Computational Linguistics, School of EECS, Peking University}\\
\affaddr{School of Foreign Languages, Peking University}\\
\email{\{linjunyang, xusun, shumingma, sukia\}@pku.edu.cn}\\
}
\date{}

\begin{document}
\maketitle
\begin{CJK}{UTF8}{gbsn}
\begin{abstract}
  In neural abstractive summarization, the conventional sequence-to-sequence (seq2seq) model often suffers from repetition and semantic irrelevance. To tackle the problem, we propose a global encoding framework, which controls the information flow from the encoder to the decoder based on the global information of the source context. It consists of a convolutional gated unit to perform global encoding to improve the representations of the source-side information. Evaluations on the LCSTS and the English Gigaword both demonstrate that our model outperforms the baseline models, and the analysis shows that our model is capable of generating summary of higher quality and reducing repetition\footnote{The code is available at \url{https://www.github.com/lancopku/Global-Encoding}}.
\end{abstract}

\section{Introduction}

Abstractive summarization can be regarded as a sequence mapping task that the source text should be mapped to the target summary. Therefore, sequence-to-sequence learning can be applied to neural abstractive summarization \citep{DBLP:conf/emnlp/KalchbrennerB13, DBLP:conf/nips/SutskeverVL14, DBLP:conf/emnlp/ChoMGBBSB14}, whose model consists of an encoder and a decoder. Attention mechanism has been broadly used in seq2seq models where the decoder extracts information from the encoder based on the attention scores on the source-side information \citep{DBLP:journals/corr/BahdanauCB14,DBLP:conf/emnlp/LuongPM15}. Many attention-based seq2seq models have been proposed  for abstractive summarization \citep{DBLP:conf/emnlp/RushCW15,Chopra2016Abstractive,DBLP:conf/conll/NallapatiZSGX16}, which outperformed the conventional statistical methods.

\begin{table}[t]
\small
\setlength{\tabcolsep}{3pt}
\centering
    \begin{tabular}{p{7.5cm}}
    \hline
    \textbf{Text: } the mainstream fatah movement on monday officially chose mahmoud abbas, chairman of the palestine liberation organization (plo), as its candidate to run for the presidential election due on jan. \#, \#\#\#\#, the official wafa news agency reported.\\
    \hline
    \textbf{seq2seq:} fatah \colorbox[rgb]{0.99,0.86,0.86}{officially officially} elects abbas as \colorbox[rgb]{0.99,0.86,0.86}{candidate} for \colorbox[rgb]{0.99,0.86,0.86}{candidate}.\\
    \hline
    \textbf{Gold:} fatah officially elects abbas as candidate for presidential election
\\
    \hline  
    \end{tabular}
    \caption{An example of the summary of the conventional attention-based seq2seq model on the Gigaword dataset. The text highlighted indicates repetition, ``\#'' refers to masked number.}
    \label{example}
\end{table}

However, recent studies show that there are salient problems in the attention mechanism. \citet{selective} pointed out that there is no obvious alignment relationship between the source text and the target summary, and the encoder outputs contain noise for the attention. For example, in the summary generated by the seq2seq in Table \ref{example}, ``officially'' is followed by the same word, as the attention mechanism still attends to the word with high attention score. Attention-based seq2seq model for abstractive summarization can suffer from repetition and semantic irrelevance, causing grammatical errors and insufficient reflection of the main idea of the source text.

To tackle this problem, we propose a model of global encoding for abstractive summarization. We set a convolutional gated unit to perform global encoding on the source context. The gate based on convolutional neural network (CNN) filters each encoder output based on the global context due to the parameter sharing, so that the representations at each time step are refined with consideration of the global context. We conduct experiments on LCSTS and Gigaword, two benchmark datasets for sentence summarization, which shows that our model outperforms the state-of-the-art methods with ROUGE-2 F1 score 26.8 and 17.8 respectively. Moreover, the analysis shows that our model is capable of reducing repetition compared with the seq2seq model.

\section{Global Encoding}

\begin{figure}
     \centering
     \includegraphics[height=6.0cm]{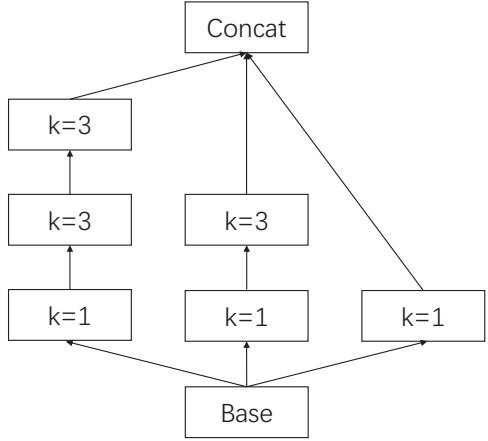}
     \caption{\textbf{Structure of our proposed Convolutional Gated Unit.} We implement 1-dimensional convolution with a structure similar to the Inception \citep{inception} over the outputs of the RNN encoder, where $k$ refers to the kernel size.} 
     \label{fig1}
\end{figure}

Our model is based on the seq2seq model with attention. For the encoder, we set a convolutional gated unit for global encoding. Based on the outputs from the RNN encoder, the global encoding refines the representation of the source context with a CNN to improve the connection of the word representation with the global context. In the following, the techniques are introduced in detail .

\subsection{Attention-based seq2seq}
The RNN encoder receives the word embedding of each word from the source text sequentially. The final hidden state with the information of the whole source text becomes the initial hidden state of the decoder. Here our encoder is a bidirectional LSTM encoder, where the encoder outputs from both directions at each time step are concatenated ($h_i\!=\![\overrightarrow{h_i}; \overleftarrow{h_i}]$).

We implement a unidirectional LSTM decoder to read the input words and generate summary word by word, with a fixed target vocabulary embedded in a high-dimensional space $Y \in R^{\mid Y \mid \times dim}$. At each time step, the decoder generates a summary word $y_{t}$ by sampling from a distribution of the target vocabulary $P_{vocab}$ until sampling the token representing the end of sentence. The hidden state of the decoder $s_{t}$ and the encoders output $h_{i}$ at each time step $i$ of the encoding process are computed with a weight matrix $W_{a}$ to obtain the global attention $\alpha_{t,i}$ and the context vector $c_{t}$. It is described below:
\begin{align}
P_{vocab} &= softmax(g([c_{t};s_{t}]))\\
s_{t} &= LSTM(y_{t-1}, s_{t-1}, C_{t-1}) \label{eq12}\\
c_{t} &= \sum^{n}_{i=1} \alpha_{t,i}h_{i} \label{eq13}\\
\alpha_{t,i} &= \frac{exp(e_{t,i})}{\sum_{j=1}^{n}exp(e_{t,j})} \label{eq14}\\
e_{t,i} &= s_{t-1}^{\top}W_{a}h_{i} \label{eq15}
\end{align}
where $C$ refers to the cell state in the LSTM, and $g(\cdot)$ refers to a non-linear function.

\subsection{Convolutional Gated Unit}

Abstractive summarization requires the core information at each encoding time step. To reach this goal, we implement a gated unit on top of the encoder outputs at each time step, which is a CNN that convolves all the encoder outputs. The parameter sharing of the convolutional kernels enables the model to extract certain types of features, specifically n-gram features. Similar to image, language also contains local correlation, such as the internal correlation of phrase structure. The convolutional units can extract these common features in the sentence and indicate the correlation among the source annotations. Moreover, to further strengthen the global information, we implement self-attention \citep{transformer} to mine the relationship of the annotation at a certain time step with other annotations. Therefore, the gated unit is able to find out both common n-gram features and global correlation. Based on the convolution and self-attention, the gated unit sets a gate to filter the source annotations from the RNN encoder, in order to select information relevant to the global semantic meaning. The global encoding allows the encoder output at each time step to become new representation vector with further connection to the global source side information. 
For convolution, we implement a structure similar to inception \citep{inception}. We use 1-dimension convolution to extract n-gram features. Following the design principle of inception, we did not use kernel where $k=5$ but instead used two kernels where $k=3$ to avoid large kernel size. The details of convolution block is described below:
\begin{align}
g_{i} &= ReLU(W[h_{i-k/2},...,h_{i+k/2}] + b)
\end{align}
where $ReLU$ refers to the non-linear activation function Rectified Linear Unit \citep{ReLU}. Based on the convolution block, we implement a structure similar to inception, as shown in Figure \ref{fig1}.

On top of the new representations generated by the CNN module, we further implement self-attention upon these representations so as to dig out the global correlations. \citet{transformer} pointed out that self-attention encourages the model to learn long-term dependencies and does not create much computational complexity, so we implement its scaled dot-product attention for the connection between the annotation at each time step and the global information:
\begin{align}
Attention(Q,K,V) &= softmax(\frac{QK^{T}}{\sqrt[]{d_{k}}})V
\end{align}
where the representations, are computed through the attention mechanism with itself and packed into a matrix. To be specific, we refer $Q$ and $V$ to the representation matrix generated by the CNN module, while $K = W_{att}V$ where $W_{att}$ is a learnable matrix.

A further step is to set a gate based on the generation from the CNN and self-attention module $g$ for the source representations $h'$ from the RNN encoder, where:
\begin{align}
\tilde{h} &= h \odot \sigma(g)
\end{align}

Since the CNN module can extract n-gram features of the whole source text and self-attention learns the long-term dependencies among the components of the input source text, the gate can perform global encoding on the encoder outputs. Based on the output of the CNN and self-attention, the logistic sigmoid function outputs a vector of value between 0 and 1 at each dimension. If the value is close to 0, the gate removes most of the information at the corresponding dimension of the source representation, and if it is close to 1, it reserves most of the information.

\subsection{Training}

In the following, we introduce the datasets that we conduct experiments on as well as our experimental settings.

Given the parameters $\theta$ and source text $x$, the models generates a summary $\tilde{y}$. The learning process is to minimize the negative log-likelihood between the generated summary $\tilde{y}$ and reference $y$:
\begin{align}
\mathcal{L} &= -\frac{1}{\mathrm{N}}\sum_{n=1}^{\mathrm{N}}\sum_{t=1}^{\mathrm{T}}p(y_{t}^{(n)}|\tilde{y}_{<t}^{(n)},x^{(n)},\theta) \label{eq22}
\end{align}
where the loss function is equivalent to maximizing the conditional probability of summary $y$ given parameters $\theta$ and source sequence $x$.

\section{Experiment Setup}

In the following, we introduce the datasets that we conduct experiments on and our experiment settings as well as the baseline models that we compare with.

\subsection{Datasets}
LCSTS is a large-scale Chinese short text summarization dataset collected from Sina Weibo, a famous Chinese social media website \citep{DBLP:conf/emnlp/HuCZ15}, consisting of more than 2.4 million text-summary pairs. The original texts are shorter than 140 Chinese characters, and the summaries are created manually. We follow the previous research \citep{DBLP:conf/emnlp/HuCZ15} to split the dataset for training, validation and testing, with 2.4M sentence pairs for training, 8K for validation and 0.7K for testing.

The English Gigaword is a sentence summarization dataset based on Annotated Gigaword \citep{napoles2012annotated}, a dataset consisting of sentence pairs, which are the first sentence of the collected news articles and the corresponding headlines. We use the data preprocessed by \citet{DBLP:conf/emnlp/RushCW15} with 3.8M sentence pairs for training, 8K for validation and 2K for testing. 

\subsection{Experiment Settings}
We implement our experiments in PyTorch on an NVIDIA 1080Ti GPU. The word embedding dimension and the number of hidden units are both 512. In both experiments, the batch size is set to 64. We use Adam optimizer \citep{DBLP:journals/corr/KingmaB14} with the default setting $\alpha = 0.001, \beta_{1}=0.9$, $\beta_{2}=0.999$ and $\epsilon=1\times10^{-8}$. The learning rate is halved every epoch. Gradient clipping is applied with range [-10, 10].

Following the previous studies, we choose ROUGE score to evaluate the performance of our model \citep{Lin2003Automatic}. ROUGE score is to calculate the degree of overlapping between generated summary and reference, including the number of n-grams. F1 scores of ROUGE-1, ROUGE-2 and ROUGE-L are used as the evaluation metrics. 

\begin{table}[t]
\centering
\begin{tabular}{llll}
\hline \bf Model & \bf R-1 & \bf R-2 & \bf R-L\\ \hline
RNN  & 21.5 & 8.9 & 18.6\\
RNN-context  & 29.9 & 17.4 & 27.2\\
CopyNet  & 34.4 & 21.6 & 31.3 \\
SRB & 33.3 & 20.0 & 30.1 \\
DRGD  & 37.0 & 24.2 & 34.2\\
\hline 
seq2seq (Our impl.) & 33.8 & 23.1 & 32.5\\
\bf +CGU & \bf 39.4 & \bf 26.9 & \bf 36.5\\ \hline
\end{tabular}
\caption{\textbf{F-Score of ROUGE on LCSTS.}}
\label{table1}
\end{table}

\subsection{Baseline Models}

As we compare our results with the results of the baseline models reported in their original papers, the evaluation on the two datasets has different baselines. In the following, we introduce the baselines for LCSTS and Gigaword respectively.

Baselines for LCSTS are introduced in the following. \textbf{RNN} and \textbf{RNN-context} are the RNN-based seq2seq models \citep{DBLP:conf/emnlp/HuCZ15}, without and with attention mechanism respectively. \textbf{CopyNet} is the attention-based seq2seq model with the copy mechanism \citep{DBLP:conf/acl/GuLLL16}. \textbf{SRB} is a model that improves semantic relevance between source text and summary \citep{SRB}. \textbf{DRGD} is the conventional seq2seq with a deep recurrent generative decoder \citep{DBLP:conf/emnlp/LiLBW17}.

As to the baselines for Gigaword, \textbf{ABS} and \textbf{ABS$+$} are the models with local attention and handcrafted features \citep{DBLP:conf/emnlp/RushCW15}. \textbf{Feats} is a fully RNN seq2seq model with some specific methods to control the vocabulary size. \textbf{RAS-LSTM} and \textbf{RAS-Elman} are seq2seq models with a convolutional encoder and an LSTM decoder and an Elman RNN decoder respectively. \textbf{SEASS} is a seq2seq model with a selective gate mechanism. \textbf{DRGD} is also a baseline for Gigaword.

Results of our implementation of the conventional seq2seq model on both datasets are also used for the evaluation of the improvement of our proposed convolutional gated unit (CGU).


\section{Analysis}

In the following sections, we report the results of our experiments and analyze the performance of our model on the evaluation of repetition. Also, we provide an example to demonstrate that our model can generate summary that is more semantically consistent with the source text.

\subsection{Results}

\begin{table}[t]
\centering
\begin{tabular}{lllll}
\hline \bf Model & \bf R-1 & \bf R-2 & \bf R-L\\ \hline
ABS  & 29.6 & 11.3 & 26.4\\
ABS$+$  & 29.8 & 11.9 & 27.0\\
Feats  & 32.7 & 15.6 & 30.6\\
RAS-LSTM  & 32.6 & 14.7 & 30.0\\
RAS-Elman  & 33.8 & 16.0 & 31.2\\
SEASS  & 36.2 & 17.5 & 33.6\\
DRGD  & \bf{36.3} & 17.6 & 33.6\\
\hline
seq2seq (Our impl.) & 33.6 & 16.3 & 31.3\\
\bf +CGU & \bf{36.3} & \bf 18.0 & \bf 33.8\\ \hline
\end{tabular}
\caption{\textbf{F-Score of ROUGE on Gigaword.}}
\label{table2}
\end{table}

In the experiments on the two datasets, our model achieves advantages of ROUGE score over the baselines, and the advantages of ROUGE score on the LCSTS are significant.
Table \ref{table1} presents the results of our model and the baselines on the LCSTS, and Table \ref{table1} shows the results of models on the Gigaword. We compare the F1 scores of our model with those of the baseline models (reported in their original articles) and our own implementation of the attention-based seq2seq. Compared with the conventional seq2seq model, our model owns an advantage of ROUGE-2 score 3.7 and 1.5 on the LCSTS and Gigaword respectively.

\subsection{Discussion}



We show a summary generated by our model, compared with that of the baseline seq2seq model and the reference. The source text introduces a phenomenon that Starbucks, an ordinary coffee brand in the United States, becomes a brand of high class and sells coffee in a much higher price. It is apparent that the main idea of the text is about the high price of Starbucks coffee in China. However, the seq2seq model generates a summary which only contains the information of the brand and the country. In addition, it has committed a mistake of redundant repetition of the word ``China''. It is not semantically relevant to the source text and it is not coherent and adequate. Compared with it, the summary of our model is more coherent and more semantically relevant to the source text. Our model focuses on the information about price instead of country, and points out the price gap in its generated summary. As ``China'' appears twice in the source text and it is hard for the baseline model to put it in a less significant place, but for our model with CGU, it is able to filter the trivial details that are irrelevant to the core meaning of the source text and just focuses on the information that contributes most to the main idea.

As our CGU is responsible for selecting important information of the outputs from the RNN encoder to improve the quality of the attention score, it should be able to reduce repetition in the generated summary. We evaluate the degree of repetition by calculating the percentage of the duplicates at the sentence level. The evaluations on the Gigaword for duplicates of 1-gram to 4 gram prove that our model significantly reduces repetition compared to the conventional seq2seq and its repetition rate is similar to the reference's. This also shows that our model is able to generate summaries of higher diversity with less repetition.

\begin{table}[t]
\small
\setlength{\tabcolsep}{3pt}
\centering
    \begin{tabular}{p{7.5 cm}}
    \hline
    \textbf{Source:} 较早进入中国市场的星巴克， 是不少小资钟情的品牌。相比在美国的平民形象，星巴克在中国就显得“高端”得多。用料并无差别的一杯中杯美式咖啡，在美国仅约合人民币12元，国内要卖21元，相当于贵了75\%。第一财经日报\\
    Starbucks, which entered Chinese market early, is a brand appealing to young people of petit bourgeoisie. Compared with its ordinary image in the United States, Starbucks seems to be of higher class in China. A Tall Americano sells about 12RMB in the United States, but 21RMB in China, which means it is 75\% more expensive.\\
    \hline
    \textbf{Reference:} 媒体称星巴克美式咖啡售价中国比美国贵75\%。\\
    Media report that the price of Starbucks Americano in China is 75\% more expensive than that in the United States.\\
    \hline
    \textbf{seq2seq:} 星巴克中国美式咖啡在中国。\\
    Starbucks China Americano in China.\\
    \hline
    \textbf{+CGU:} 星巴克美式咖啡中国贵75\%。\\
    Starbucks Americano is 75\% more expensive in China.\\
    \hline
    \end{tabular}
    \caption{An example of our summarization, compared with that of the seq2seq model and the reference.}
    \label{sum}
\end{table}



\section{Related Work}

Researchers developed many statistical methods and linguistic-rule-based methods to study automatic summarization \citep{banko2000headline,dorr2003hedge,zajic2004bbn,cohn2008sentence}. With the development of Neural Network in NLP, more and more researches have appeared in abstractive summarization since it seems possible that Neural Network can help achieve the two goals. \citet{DBLP:conf/emnlp/RushCW15} first applied sequence-to-sequence model with attention mechanism to abstractive summarization and realized significant achievements. \citet{Chopra2016Abstractive} changed the ABS model with an RNN decoder and \citet{DBLP:conf/conll/NallapatiZSGX16} changed the system to a fully-RNN sequence-to-sequence model and achieved outstanding performance. \citet{selective} proposed a selective gate mechanism to filter secondary information. \citet{DBLP:conf/emnlp/LiLBW17} proposed a deep recurrent generative decoder to learn latent structure information. \citet{wean} proposed a model that generates words by querying word embeddings.

\section{Conclusion}

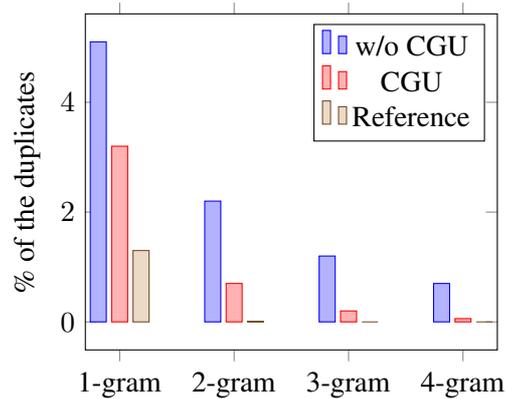
\begin{figure}[t]
\begin{tikzpicture}
\begin{axis}[legend pos=north east, ybar, symbolic x coords={1-gram, 2-gram, 3-gram, 4-gram}, ylabel = {\% of the duplicates}, xtick=data, bar width=6pt]
\addlegendentry{w/o CGU}
\addplot 
coordinates{(1-gram, 5.1)(2-gram, 2.2)(3-gram, 1.2)(4-gram, 0.7)};

\addlegendentry{CGU}
\addplot 
coordinates{(1-gram, 3.2)(2-gram, 0.7)(3-gram, 0.2)(4-gram, 0.06)};

\addlegendentry{Reference}
\addplot 
coordinates{(1-gram, 1.3)(2-gram, 0.01)(3-gram, 0)(4-gram, 0)};

\end{axis}
\end{tikzpicture}
\caption{\textbf{Percentage of the duplicates at sentence level.} Evaluated on the Gigaword. }
\end{figure}

In this paper, we propose a new model for abstractive summarization. The convolutional gated unit performs global encoding on the source side information so that the core information can be reserved and the secondary information can be filtered. Experiments on the LCSTS and Gigaword show that our model outperforms the baselines, and the analysis shows that it is able to reduce repetition in the generated summaries, and it is more robust to inputs of different lengths, compared with the conventional seq2seq model. 

\section*{Acknowledgements}
This work was supported in part by National Natural Science Foundation of China (No. 61673028), National High Technology Research and Development Program of China (863 Program, No. 2015AA015404), and the National Thousand Young Talents Program. Xu Sun is the corresponding author of this paper.

\nocite{SunWei2017,amr}

\end{CJK}

\bibliography{acl2018}

\begin{thebibliography}{26}
\expandafter\ifx\csname natexlab\endcsname\relax\def\natexlab#1{#1}\fi

\bibitem[{Bahdanau et~al.(2014)Bahdanau, Cho, and
  Bengio}]{DBLP:journals/corr/BahdanauCB14}
Dzmitry Bahdanau, Kyunghyun Cho, and Yoshua Bengio. 2014.
\newblock Neural machine translation by jointly learning to align and
  translate.
\newblock \emph{CoRR}, abs/1409.0473.

\bibitem[{Banko et~al.(2000)Banko, Mittal, and Witbrock}]{banko2000headline}
Michele Banko, Vibhu~O Mittal, and Michael~J Witbrock. 2000.
\newblock Headline generation based on statistical translation.
\newblock In \emph{Proceedings of the 38th Annual Meeting on Association for
  Computational Linguistics}, pages 318--325. Association for Computational
  Linguistics.

\bibitem[{Cho et~al.(2014)Cho, van Merrienboer, G{\"{u}}l{\c{c}}ehre, Bahdanau,
  Bougares, Schwenk, and Bengio}]{DBLP:conf/emnlp/ChoMGBBSB14}
Kyunghyun Cho, Bart van Merrienboer, {\c{C}}aglar G{\"{u}}l{\c{c}}ehre, Dzmitry
  Bahdanau, Fethi Bougares, Holger Schwenk, and Yoshua Bengio. 2014.
\newblock Learning phrase representations using {RNN} encoder-decoder for
  statistical machine translation.
\newblock In \emph{{EMNLP} 2014}, pages 1724--1734.

\bibitem[{Chopra et~al.(2016)Chopra, Auli, and Rush}]{Chopra2016Abstractive}
Sumit Chopra, Michael Auli, and Alexander~M. Rush. 2016.
\newblock Abstractive sentence summarization with attentive recurrent neural
  networks.
\newblock In \emph{Conference of the North American Chapter of the Association
  for Computational Linguistics: Human Language Technologies}, pages 93--98.

\bibitem[{Cohn and Lapata(2008)}]{cohn2008sentence}
Trevor Cohn and Mirella Lapata. 2008.
\newblock Sentence compression beyond word deletion.
\newblock In \emph{Proceedings of the 22nd International Conference on
  Computational Linguistics-Volume 1}, pages 137--144. Association for
  Computational Linguistics.

\bibitem[{Dorr et~al.(2003)Dorr, Zajic, and Schwartz}]{dorr2003hedge}
Bonnie Dorr, David Zajic, and Richard Schwartz. 2003.
\newblock Hedge trimmer: A parse-and-trim approach to headline generation.
\newblock In \emph{Proceedings of the HLT-NAACL 03 on Text summarization
  workshop-Volume 5}, pages 1--8. Association for Computational Linguistics.

\bibitem[{Gu et~al.(2016)Gu, Lu, Li, and Li}]{DBLP:conf/acl/GuLLL16}
Jiatao Gu, Zhengdong Lu, Hang Li, and Victor O.~K. Li. 2016.
\newblock Incorporating copying mechanism in sequence-to-sequence learning.
\newblock In \emph{{ACL} 2016}.

\bibitem[{Hu et~al.(2015)Hu, Chen, and Zhu}]{DBLP:conf/emnlp/HuCZ15}
Baotian Hu, Qingcai Chen, and Fangze Zhu. 2015.
\newblock {LCSTS:} {A} large scale chinese short text summarization dataset.
\newblock In \emph{{EMNLP} 2015}, pages 1967--1972.

\bibitem[{Kalchbrenner and Blunsom(2013)}]{DBLP:conf/emnlp/KalchbrennerB13}
Nal Kalchbrenner and Phil Blunsom. 2013.
\newblock Recurrent continuous translation models.
\newblock In \emph{{EMNLP} 2013}, pages 1700--1709.

\bibitem[{Kingma and Ba(2014)}]{DBLP:journals/corr/KingmaB14}
Diederik~P. Kingma and Jimmy Ba. 2014.
\newblock Adam: {A} method for stochastic optimization.
\newblock \emph{CoRR}, abs/1412.6980.

\bibitem[{Li et~al.(2017)Li, Lam, Bing, and Wang}]{DBLP:conf/emnlp/LiLBW17}
Piji Li, Wai Lam, Lidong Bing, and Zihao Wang. 2017.
\newblock Deep recurrent generative decoder for abstractive text summarization.
\newblock In \emph{{EMNLP} 2017,}, pages 2091--2100.

\bibitem[{Lin and Hovy(2003)}]{Lin2003Automatic}
Chin~Yew Lin and Eduard Hovy. 2003.
\newblock Automatic evaluation of summaries using n-gram co-occurrence
  statistics.
\newblock In \emph{Conference of the North American Chapter of the Association
  for Computational Linguistics on Human Language Technology}, pages 71--78.

\bibitem[{Luong et~al.(2015)Luong, Pham, and
  Manning}]{DBLP:conf/emnlp/LuongPM15}
Thang Luong, Hieu Pham, and Christopher~D. Manning. 2015.
\newblock Effective approaches to attention-based neural machine translation.
\newblock In \emph{{EMNLP} 2015}, pages 1412--1421.

\bibitem[{Ma et~al.(2018)Ma, Sun, Li, Li, Li, and Ren}]{wean}
Shuming Ma, Xu~Sun, Wei Li, Sujian Li, Wenjie Li, and Xuancheng Ren. 2018.
\newblock Query and output: Generating words by querying distributed word
  representations for paraphrase generation.
\newblock In \emph{{NAACL} 2018}.

\bibitem[{Ma et~al.(2017)Ma, Sun, Xu, Wang, Li, and Su}]{SRB}
Shuming Ma, Xu~Sun, Jingjing Xu, Houfeng Wang, Wenjie Li, and Qi~Su. 2017.
\newblock Improving semantic relevance for sequence-to-sequence learning of
  chinese social media text summarization.
\newblock In \emph{{ACL} 2017}, pages 635--640.

\bibitem[{Nair and Hinton(2010)}]{ReLU}
Vinod Nair and Geoffrey~E. Hinton. 2010.
\newblock Rectified linear units improve restricted boltzmann machines.
\newblock In \emph{ICML 2010}, pages 807--814.

\bibitem[{Nallapati et~al.(2016)Nallapati, Zhou, dos Santos,
  G{\"{u}}l{\c{c}}ehre, and Xiang}]{DBLP:conf/conll/NallapatiZSGX16}
Ramesh Nallapati, Bowen Zhou, C{\'{\i}}cero~Nogueira dos Santos, {\c{C}}aglar
  G{\"{u}}l{\c{c}}ehre, and Bing Xiang. 2016.
\newblock Abstractive text summarization using sequence-to-sequence rnns and
  beyond.
\newblock In \emph{CoNLL 2016}, pages 280--290.

\bibitem[{Napoles et~al.(2012)Napoles, Gormley, and
  Van~Durme}]{napoles2012annotated}
Courtney Napoles, Matthew Gormley, and Benjamin Van~Durme. 2012.
\newblock Annotated gigaword.
\newblock In \emph{Proceedings of the Joint Workshop on Automatic Knowledge
  Base Construction and Web-scale Knowledge Extraction}, pages 95--100.
  Association for Computational Linguistics.

\bibitem[{Rush et~al.(2015)Rush, Chopra, and Weston}]{DBLP:conf/emnlp/RushCW15}
Alexander~M. Rush, Sumit Chopra, and Jason Weston. 2015.
\newblock A neural attention model for abstractive sentence summarization.
\newblock In \emph{{EMNLP} 2015}, pages 379--389.

\bibitem[{Sun et~al.(2017)Sun, Wei, Ren, and Ma}]{SunWei2017}
Xu~Sun, Bingzhen Wei, Xuancheng Ren, and Shuming Ma. 2017.
\newblock Label embedding network: Learning label representation for soft
  training of deep networks.
\newblock \emph{CoRR}, abs/1710.10393.

\bibitem[{Sutskever et~al.(2014)Sutskever, Vinyals, and
  Le}]{DBLP:conf/nips/SutskeverVL14}
Ilya Sutskever, Oriol Vinyals, and Quoc~V. Le. 2014.
\newblock Sequence to sequence learning with neural networks.
\newblock In \emph{Advances in Neural Information Processing Systems 27: Annual
  Conference on Neural Information Processing Systems 2014}, pages 3104--3112.

\bibitem[{Szegedy et~al.(2015)Szegedy, Vanhoucke, Ioffe, Shlens, and
  Wojna}]{inception}
Christian Szegedy, Vincent Vanhoucke, Sergey Ioffe, Jonathon Shlens, and
  Zbigniew Wojna. 2015.
\newblock Rethinking the inception architecture for computer vision.
\newblock \emph{CoRR}, abs/1512.00567.

\bibitem[{Takase et~al.(2016)Takase, Suzuki, Okazaki, Hirao, and Nagata}]{amr}
Sho Takase, Jun Suzuki, Naoaki Okazaki, Tsutomu Hirao, and Masaaki Nagata.
  2016.
\newblock Neural headline generation on abstract meaning representation.
\newblock In \emph{{EMNLP} 2016}, pages 1054--1059.

\bibitem[{Vaswani et~al.(2017)Vaswani, Shazeer, Parmar, Uszkoreit, Jones,
  Gomez, Kaiser, and Polosukhin}]{transformer}
Ashish Vaswani, Noam Shazeer, Niki Parmar, Jakob Uszkoreit, Llion Jones,
  Aidan~N. Gomez, Lukasz Kaiser, and Illia Polosukhin. 2017.
\newblock Attention is all you need.
\newblock In \emph{NIPS 2017}, pages 6000--6010.

\bibitem[{Zajic et~al.(2004)Zajic, Dorr, and Schwartz}]{zajic2004bbn}
David Zajic, Bonnie Dorr, and Richard Schwartz. 2004.
\newblock Bbn/umd at duc-2004: Topiary.
\newblock In \emph{Proceedings of the HLT-NAACL 2004 Document Understanding
  Workshop, Boston}, pages 112--119.

\bibitem[{Zhou et~al.(2017)Zhou, Yang, Wei, and Zhou}]{selective}
Qingyu Zhou, Nan Yang, Furu Wei, and Ming Zhou. 2017.
\newblock Selective encoding for abstractive sentence summarization.
\newblock In \emph{{ACL} 2017}, pages 1095--1104.

\end{thebibliography}
\bibliographystyle{acl_natbib}

\end{document}